\documentclass[10pt,twocolumn,letterpaper]{article}
\usepackage[accsupp]{axessibility}
\usepackage[final]{cvpr}

\definecolor{cvprblue}{rgb}{0.21,0.49,0.74}
\usepackage[pagebackref,breaklinks,colorlinks,allcolors=cvprblue]{hyperref}

\def\paperID{V4A-34} 
\def\confName{CVPR}
\def\confYear{2025}

\title{California Crop Yield Benchmark: Combining Satellite Image, Climate, Evapotranspiration, and Soil Data Layers for County-Level Yield Forecasting of Over 70 Crops}

\author{
Hamid Kamangir\textsuperscript{1}\qquad
Mona Hajiesmaeeli\textsuperscript{2}\qquad
Mason Earles\textsuperscript{1,3}\\[1ex]
\textsuperscript{1}University of California, Davis\\
\textsuperscript{2} Texas A\&M University-Corpus Christi\\
\textsuperscript{3}AI Institute for Food Systems\\[1ex]
\texttt{\small hkamangir@ucdavis.edu, mona.hajiesmaeeli@tamucc.ed, jmearles@ucdavis.edu}
}

\begin{document}
\maketitle
\begin{abstract}
California is a global leader in agricultural production, contributing 12.5\% of the United States’ total output and ranking as the fifth-largest food and cotton supplier in the world. Despite the availability of extensive historical yield data from the USDA National Agricultural Statistics Service, accurate and timely crop yield forecasting remains a challenge due to the complex interplay of environmental, climatic, and soil-related factors. In this study, we introduce a comprehensive crop yield benchmark dataset covering over 70 crops across all California counties from 2008 to 2022. The benchmark integrates diverse data sources, including Landsat satellite imagery, daily climate records, monthly evapotranspiration, and high-resolution soil properties.
To effectively learn from these heterogeneous inputs, we develop a multi-modal deep learning model tailored for county-level, crop-specific yield forecasting. The model employs stratified feature extraction and a time-series encoder to capture spatial and temporal dynamics during the growing season. Static inputs such as soil characteristics and crop identity inform long-term variability.
Our approach achieves an overall $R^2$ score of 0.76 across all crops of unseen test dataset, highlighting strong predictive performance across California’s diverse agricultural regions. This benchmark and modeling framework offer a valuable foundation for advancing agricultural forecasting, climate adaptation, and precision farming.
The full dataset and codebase are publicly available at our \href{https://github.com/plant-ai-biophysics-lab/california-crop-yield-benchmark}{GitHub repository}.
\end{abstract}    
\section{Introduction}
\label{sec:intro}
California is renowned for its diverse agricultural production, encompassing grains, fruits, vegetables, legumes, forage, and industrial crops.
\begin{figure}[ht]
\centering
\captionsetup{font=footnotesize}%
\includegraphics[width=8cm, height=7cm]{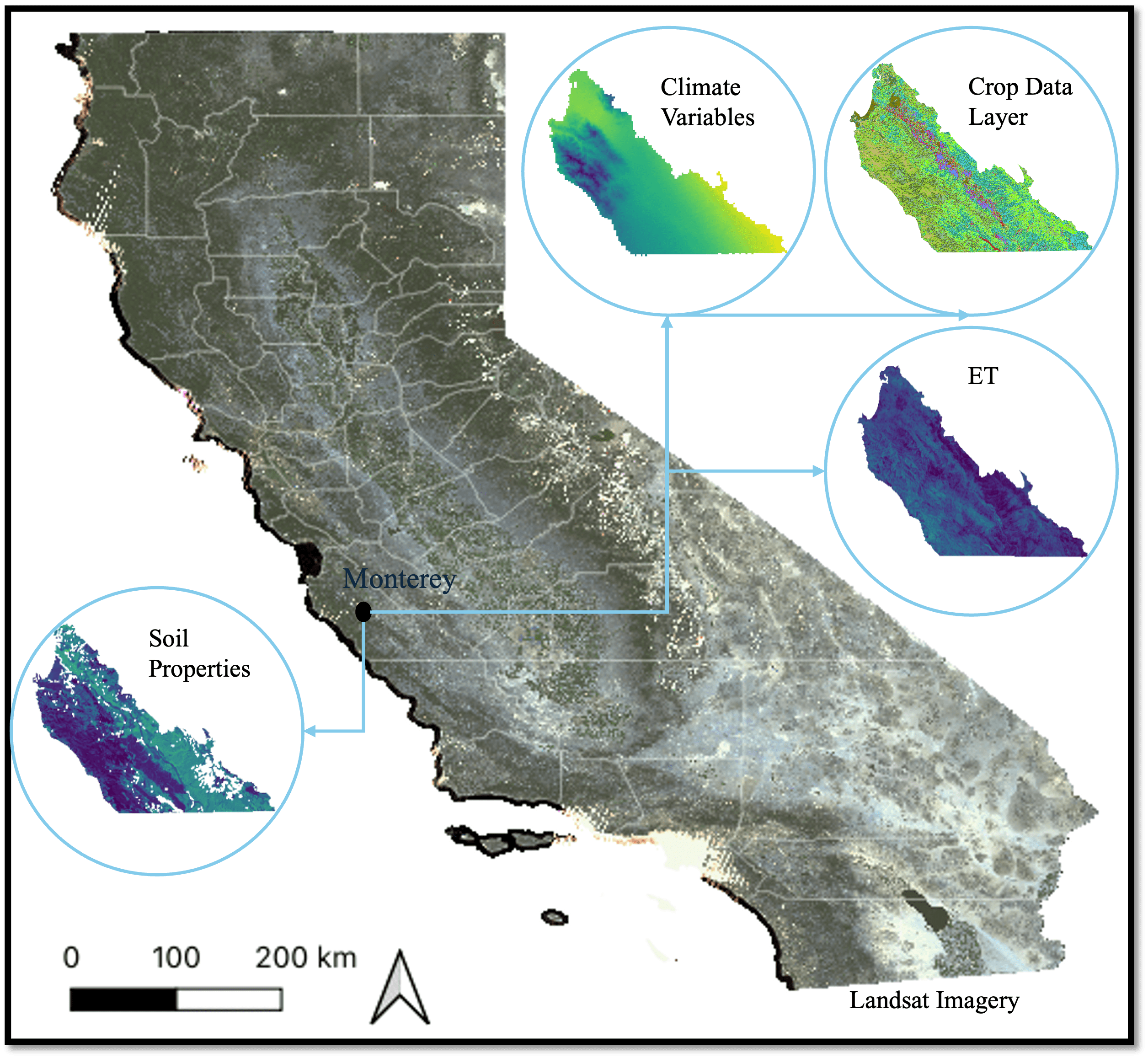} 
\caption{Sample visualization of California Landsat imagery highlighting Monterey County, overlaid with evapotranspiration (ET), climate, and soil data. This figure represents a subset of a broader dataset collected across all counties in California from 2008 to 2022 for large-scale crop yield forecasting.
 }
\label{fig:init}
\end{figure}

 As a leader in the industry, the state prioritizes high-quality cultivation across its vast agricultural landscape. Accounting for 12.5\% of all agricultural production in the United States, California plays a crucial role in the nation’s food supply. Remarkably, it stands as the world’s fifth-largest supplier of food and cotton, reinforcing its significance in global agriculture \citep{CDFA_Statistics}.

A key resource for understanding and predicting crop yields is the United States Department of Agriculture (USDA) National Agricultural Statistics Service (NASS) California Field Office, which provides annual county-level yield reports for over 100 different crops \citep{USDA_NASS}. This extensive dataset, built upon years of historical observations, serves as an invaluable foundation for yield forecasting. The long-term records offer crucial insights into trends, anomalies, and the impact of environmental variability, while the diversity of crops ensures broad applicability across different agricultural systems. Leveraging this dataset enhances predictive modeling efforts, providing a data-rich baseline for integrating meteorological, soil, and remote sensing information into crop yield predictions. However, accurate and timely forecasting remains a complex challenge, as it requires accounting for multiple dynamic factors, including: short- and long-term weather variations influenced by meteorological conditions during the growing season,  soil water consumption, soil nutrient availability, and high-resolution remote sensing data.

While process-based prediction approaches \citep{tao2009modelling, chenu2017contribution, jones2017toward, xu2019global} exist, they often suffer from inaccuracies due to strong assumptions about management practices \citep{fan2022gnn, kamangir2024cmavit}. In contrast, deep learning (DL)-based methods, inspired by the success of deep neural networks \citep{krizhevsky2012imagenet}, have gained widespread adoption for crop yield prediction \citep{khaki2021simultaneous, cheng2022high, kamangir2024large, kamangir2024cmavit}. These methods demonstrate effectiveness in precise agricultural tracking \citep{khaki2021simultaneous, kamangir2024large, kamangir2024cmavit} and capture the spatial and temporal variability of meteorological data \citep{fan2022gnn, kamangir2024large}. The growing availability of multi-source remote sensing, climate reanalysis, and machine learning creates opportunities for large-scale, high-resolution yield prediction. However, integrating diverse data modalities—such as satellite imagery, climate variables, soil properties, and evapotranspiration—remains a challenge. Existing datasets often lack standardization and spatial coverage at the county level across California’s full range of crops. Most prior studies have focused on one or two major crops, limiting their generalizability. Addressing these issues is essential for improving the robustness of yield forecasting models across diverse environments.

Deep learning-based crop yield prediction methods generally fall into two categories: remote sensing-driven and meteorological data-driven approaches. The first category \citep{khaki2021simultaneous, wu2021spatiotemporal, falco2021influence, garnot2021panoptic, cheng2022high, kamangir2024large} relies on satellite imagery, UAV-captured images, and vegetation indices to estimate annual yields. The second category \citep{gandhi2016rice, akhavizadegan2021time, mourtzinis2021advancing} uses meteorological parameters—such as temperature, precipitation, and vapor pressure deficit—for yield forecasting. However, remote sensing methods may miss the direct influence of climate on crop growth, while meteorological approaches lack spatial resolution crucial for monitoring. Relying solely on either modality is insufficient, as factors like soil properties, evapotranspiration, and management practices are critical to crop productivity and must be integrated for more comprehensive predictions.

Transformers, originally introduced in natural language processing (NLP) \citep{vaswani2017attention}, have been adapted to computer vision and crop forecasting. The Vision Transformer (ViT) architecture \citep{dosovitskiy2020image} partitions an image into patches and applies Multi-Head Self-Attention (MHSA) \citep{vaswani2017attention} to capture global representations. Extensions such as DeiT \citep{touvron2021training} improve efficiency through knowledge distillation, Swin Transformer \citep{liu2021swin} uses shifted windows for scalability, PVT \citep{wang2021pyramid} handles dense predictions, and MAE \citep{he2022masked} enables self-supervised learning. Despite these advancements, applying standard ViT models to crop forecasting poses challenges. Yield prediction relies on multi-modal inputs—not just images—including climate, soil, and agronomic data. Traditional ViTs, designed for single-modality tasks, struggle with integrating heterogeneous sources, modeling high-resolution spatial dependencies, and capturing long-term temporal variation—all vital in agricultural prediction.

Multi-modal learning has emerged as a promising approach to address these limitations. One pioneering effort is the MMST-ViT model by Lin et al. (2023) \citep{lin2023mmst}, which integrates satellite imagery, meteorological data, and climate trends. The model uses a multi-modal contrastive learning technique for pre-training, reducing reliance on labeled data. It is evaluated across over 200 U.S. counties and outperforms models like CNN-RNN and GNN-RNN. However, it targets only major crops such as corn, cotton, soybean, and winter wheat, limiting its general scope. Another approach, CMAViT, introduced by Kamangir et al. (2024) \citep{kamangir2024cmavit}, fuses Sentinel imagery, climate data, and categorical management variables to estimate vineyard yield. CMAViT demonstrates the potential of multi-modal fusion for incorporating both environmental and human factors in forecasting. However, it is constrained to a small geographic area and focuses solely on vineyards, which limits its generalizability to broader agricultural systems. Together, these studies highlight the promise and ongoing challenges of applying deep learning to crop yield forecasting. Integrating multi-modal data sources remains essential for improving prediction accuracy and ensuring models are robust across crop types, regions, and management conditions.

In contrast, our work advances multi-modal learning for crop yield forecasting by incorporating a more diverse set of modalities, including remote sensing, climate data, soil properties, and evapotranspiration (ET). Furthermore, our model extends its applicability to 70 different crops and covers all counties in California, providing a more comprehensive and scalable approach to yield prediction across diverse agricultural landscapes.

To tackle these challenges, this study has two primary objectives: 
\begin{enumerate}
    \item Developing a large-scale dataset that integrates Landsat satellite imagery, climate records, soil properties, and evapotranspiration (ET) data at a spatial resolution of $30m$ for remote sensing data and $1000m$ for climate variables, covering all counties in California. This dataset is specifically designed for county-level crop yield forecasting of 70 different crops, ensuring comprehensive and standardized inputs for model training and evaluation.
    \item Developing a novel multi-modal deep learning model that processes multi-source time-series data to predict crop yields throughout the growing season. The model employs a time-series multimodal encoder, allowing it to capture seasonal trends and crop growth dynamics across different environmental conditions. By incorporating static soil properties at each time step, the model enhances its ability to understand the long-term influence of soil characteristics on yield variations. 
\end{enumerate}

This research contributes both a high-quality benchmark dataset and a state-of-the-art machine learning framework for large-scale crop yield prediction at the county level, providing a scalable approach for agricultural forecasting.

\begin{figure*}[ht]
\centering
\captionsetup{font=footnotesize}%
\includegraphics[width=16cm, height=13cm]{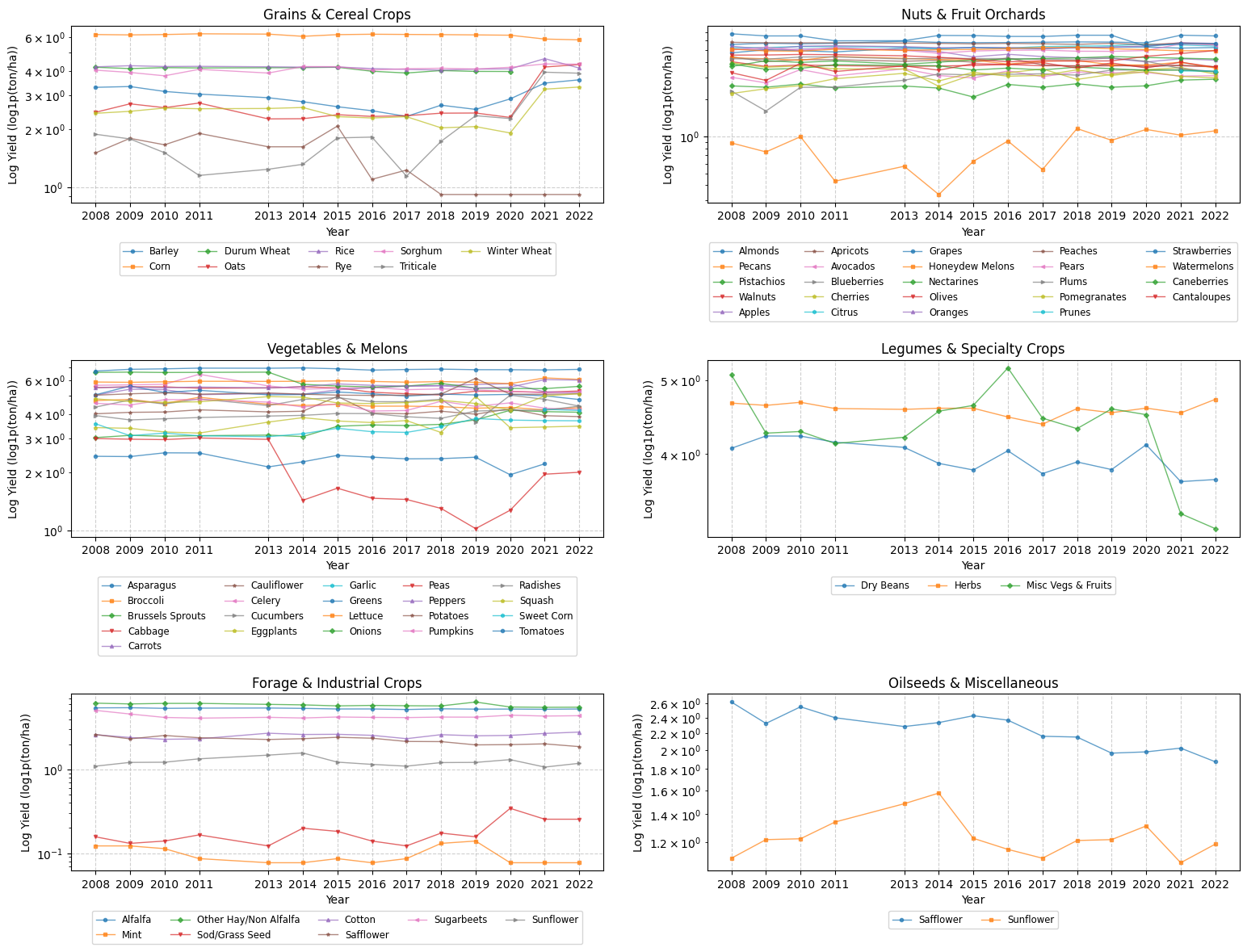} 
\caption{Yield trends from 2008 to 2022 for 70 crop types across all counties in California, grouped by crop category: Grains \& Cereal Crops, Nuts \& Fruit Orchards, Vegetables \& Melons, Legumes \& Specialty Crops, Forage \& Industrial Crops, and Oilseeds \& Miscellaneous. Yields are shown on a logarithmic scale (log10 tons per hectare) to capture the variability and distribution across diverse crop types.
 }
\label{fig:crop yield}
\end{figure*}

\begin{table*}[h]
    \centering
    \footnotesize
    \begin{tabular}{p{1cm} p{3.8cm} p{1cm} p{3cm} p{1.3cm} p{1.3cm} p{1.3cm} p{1.5cm}} 
        \toprule
        \textbf{Category} & \textbf{Variables} & \textbf{Unit} & \textbf{Related Properties} & \textbf{Spatial Resolution} & \textbf{Temporal Resolution} & \textbf{Year} & \textbf{Source} \\ 
        \bottomrule
        Yield & Crop Yield  & ton/ha & Ground-truth & N/A & Yearly & 2008-2022 & USDA \cite{USDA_CroplandCROS} \\  
        \midrule
        Satellite & RGB, NIR, SWIR-1, SWIR-2 & Reflectance & Plant vigor & 30 m & Monthly & 2008-2022 & Landsat \\  
        \midrule
        \multirow{8}{*}{Climate} 
        & Minimum Temperature (tmin) & °C & Temperature variability & \multirow{8}{*}{1000 m} & \multirow{8}{*}{Daily} & \multirow{8}{*}{2008-2022} & \multirow{8}{*}{Daymet \citep{thornton2014daymet}} \\  
        & Maximum Temperature (tmax) & °C & Temperature variability &  &  &  & \\  
        & Precipitation (prcp) & mm & Water availability &  &  &  & \\  
        & Daylight (dayl) & seconds & Solar exposure &  &  &  & \\  
        & Solar Radiation (srad) & W/m² & Energy balance &  &  &  & \\  
        & Vapor Pressure (vp) & kPa & Atmospheric moisture &  &  &  & \\  
        & Snow Water Equivalent (snow) & mm & Snowpack assessment &  &  &  & \\  
        & Potential Evapotranspiration (pet) & mm & Water demand &  &  &  & \\  
        \midrule 
        \multirow{5}{*}{Soil} 
        & Available Water Storage & cm & Soil moisture retention & \multirow{5}{*}{30 m} & \multirow{5}{*}{Static} & \multirow{5}{*}{2023} & \multirow{5}{*}{SSURGO \citep{SSURGO}} \\  
        & Slope Gradient & \% & Terrain steepness &  &  &  & \\  
        & Water Supply & mm & Soil water availability &  &  &  & \\  
        & Drainage Class & Categorical & Soil drainage characteristics &  &  &  & \\  
        & Hydrologic Soil Group & Categorical & Runoff potential &  &  &  & \\  
        \midrule
        ET & Evapotranspiration & mm & Water loss through soil & 30 m & Monthly & 2008-2022 & OpenET \citep{melton2022openet} \\  
        \bottomrule
    \end{tabular}
    \caption{Data sources and their associated parameters.}
    \label{tab:data_sources}
\end{table*}

\section{Material}
\subsection{Study area and crop yield data}
The study focuses on mapping 70 different crops across all counties in California, U.S. These crops represent major agricultural commodities in the region, with substantial harvested area and yield observation data available. The primary data source for yield measurements is the USDA County-Based Yield Reports, which provide annual crop yield estimates at the county level \citep{USDA_NASS}. These reports serve as a high-quality ground-truth dataset, offering detailed insights into crop production trends over time.

The dataset covers yield observations from 2008 to 2022, except for 2012, capturing temporal variations in agricultural productivity. Given the significance of California’s agricultural economy, the selected crops span multiple categories, including grains, fruits, vegetables, legumes, forage, and industrial crops. 
Figure~\ref{fig:crop yield} illustrates the yield trends for different crop categories, showcasing the inter-annual variability in production levels across the study period. The dataset’s granularity at the county level allows for a fine-scale analysis of spatial and temporal dynamics in agricultural production.
This county-level yield observation selected as the primary ground-truth due to their consistent methodology, long-term availability, and extensive spatial coverage across California.

\subsection{California Yield Benchmark Dataset}
The CA Yield Benchmark aims to be a publicly available, large-scale crop yield dataset with high resolution and multiple variables, including multi-band Landsat imagery, climate data, evapotranspiration (ET), and soil properties, covering the years 2008 to 2022, except for 2012. The exclusion of 2012 is due to the availability of only a single Landsat satellite that year, making it challenging to obtain free cloud-free imagery for each month. The dataset is explained in detail in Table~\ref{tab:data_sources} and the following subsections. 
\label{sec:formatting}

\subsection{Cropland Data Layer}
The U.S. Department of Agriculture (USDA) National Agricultural Statistics Service (NASS) has consistently produced the annual Cropland Data Layer (CDL) for many years. This comprehensive dataset provides information on over 100 crop types as well as significant non-agricultural land use types, covering the entire Conterminous United States (CONUS) at a 30-meter spatial resolution.
For this study, we extracted the boundaries of cultivated crop fields that are at least 10 hectares in size using crop type maps from the CDL generated by NASS for years 2008 to 2022 \cite{USDA_CroplandCROS}.

\subsection{Satellite Data}
Remote sensing plays a crucial role in agricultural monitoring by providing consistent and long-term observations of land cover and crop dynamics. For this study, we utilize Landsat satellite imagery, which offers a rich historical archive dating back to 1985. However, to align with the availability of the CDL, our analysis focuses on data from 2008 onward. The year 2012 was excluded due to the temporary absence of a second operational sensor, which limited the ability to compile a cloud-free time-series dataset. We selected key spectral bands—Red, Green, Blue (RGB), Near Infrared (NIR), and Shortwave Infrared (SWIR-1 and SWIR-2)—as they provide valuable information on vegetation health, soil moisture, and land use changes. To capture seasonal variations, we acquired imagery at a monthly frequency, resulting in 12 observations per year. The data has been collected for all counties.

\subsection{Evapotranspiration Data}
Evapotranspiration (ET) is a critical variable for understanding crop water use and monitoring agricultural sustainability. Remote sensing-based ET datasets provide valuable insights into water availability and consumption across different landscapes. For this study, we utilize data from OpenET \citep{melton2022openet}, a collaborative initiative that leverages satellite observations and multiple ET models to estimate field-scale evapotranspiration across the United States. 

\begin{figure*}[ht]
\centering
\captionsetup{font=footnotesize}%
\includegraphics[width=17cm, height=6cm]{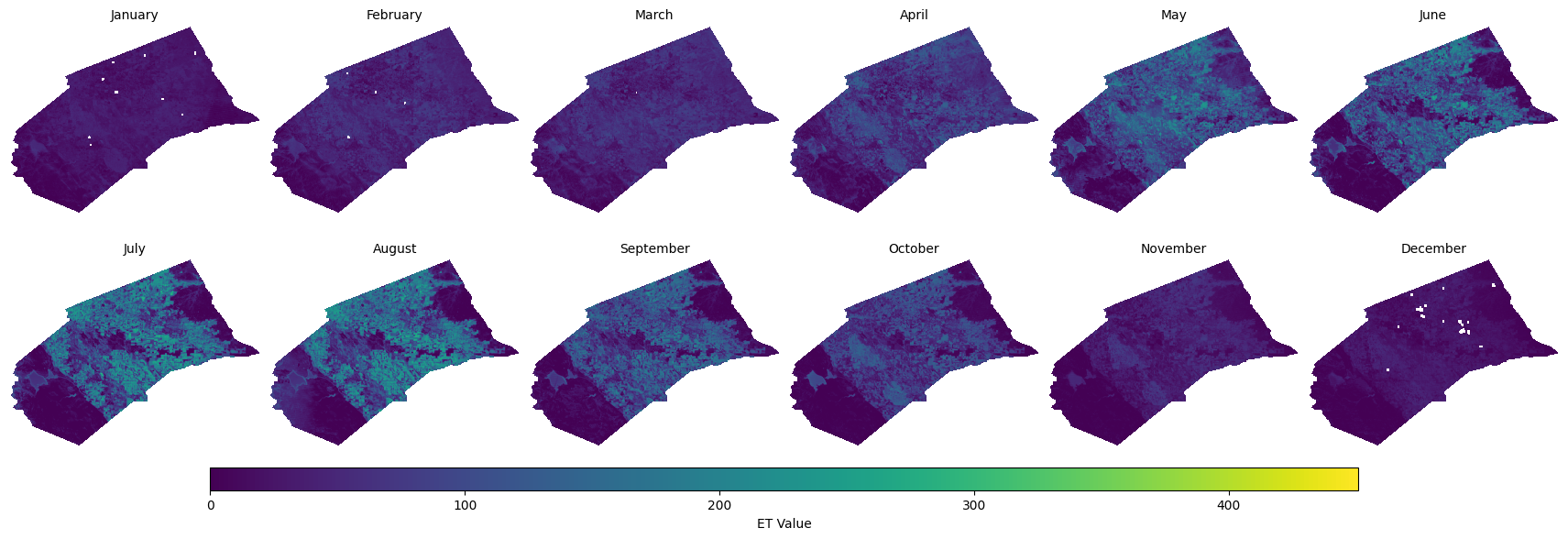} 
\caption{Sample time series visualization of the evapotranspiration (ET) variable for Merced County, California, 2022. This figure serves as an example from a statewide dataset collected across all California counties from 2008 to 2022.
 }
\label{fig:et}
\end{figure*}

OpenET primarily relies on Landsat imagery and integrates multiple approaches, including energy balance and machine learning-based models, to provide robust ET estimates at a 30-meter spatial resolution. This dataset enables us to analyze historical and recent trends in water use efficiency across cultivated croplands. To ensure consistency with the CDL dataset, we focus on OpenET data from 2008 onward, excluding 2012 due to the temporary absence of a second operational Landsat sensor, sample of timeseries ET data has been visualized in Figure~\ref{fig:et}. Monthly ET estimates are extracted to align with our temporal analysis framework, ensuring comprehensive seasonal monitoring of crop water use.

\subsection{Climate Data}
Climate variables such as temperature, precipitation, and solar radiation play a crucial role in crop growth, yield variability, and overall agricultural productivity. To incorporate these environmental factors into our analysis, we utilize climate data from DayMet, a high-resolution dataset that provides daily gridded weather observations across North America \citep{thornton2014daymet}. 

\begin{figure*}[ht]
\centering
\captionsetup{font=footnotesize}%
\includegraphics[width=17cm, height=6.5cm]{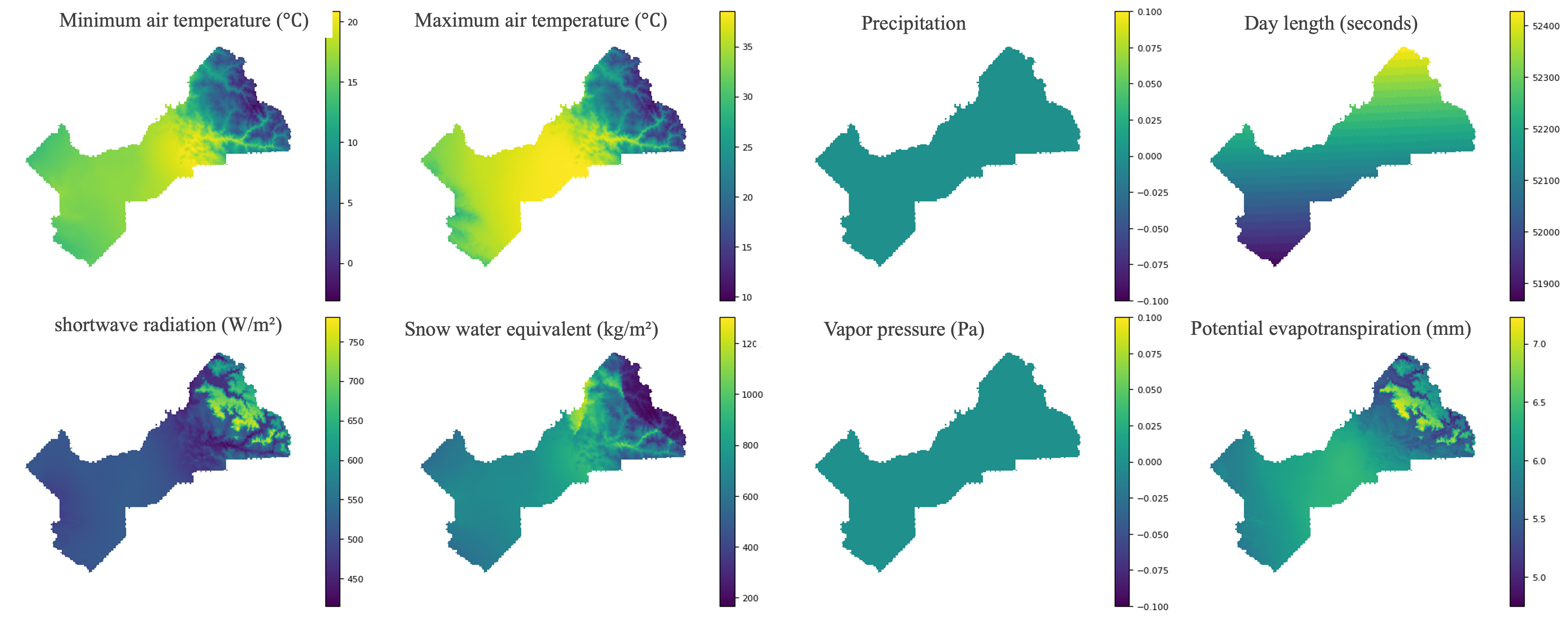} 
\caption{Sample visualization of the 8 climate variables for Fresno County, California, day 180, year 2022. This figure serves as an example from a statewide dataset collected across all California counties from 2008 to 2022.
 }
\label{fig:climate}
\end{figure*}

\begin{figure*}[ht]
\centering
\captionsetup{font=footnotesize}%
\includegraphics[width=17cm, height=3cm]{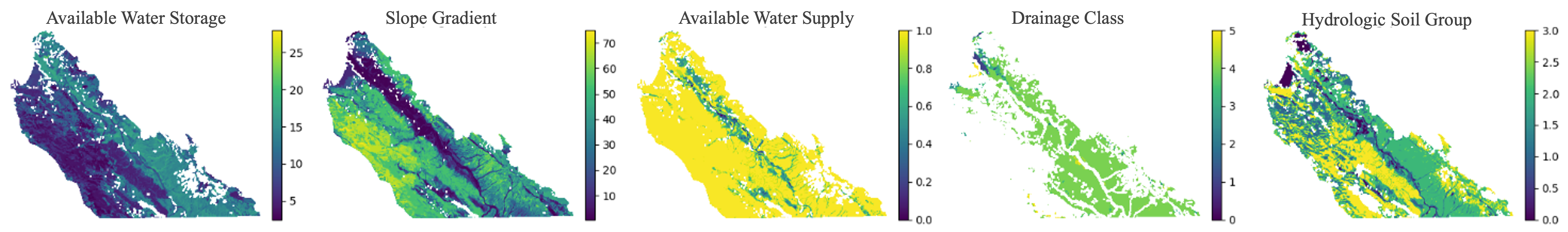} 
\caption{Sample visualization of five rasterized soil properties from the SSURGO database at 30-meter spatial resolution for Monterey County, California. This figure represents a subset of a comprehensive statewide dataset covering all California counties.
 }
\label{fig:soil}
\end{figure*}

DayMet offers spatially continuous estimates of key meteorological variables at a $1 \text{ km} \times 1 \text{ km}$ resolution, enabling detailed environmental monitoring. For this study, we extract daily climate data corresponding to the spatial extent of our cropland dataset, ensuring alignment with the CDL and OpenET datasets (sample of climate variable has been presented in Figure~\ref{fig:climate}). The following eight climate variables were selected based on their relevance to crop growth and water use:  Minimum air temperature (tmin) and maximum air temperature (tmax) (°C), crucial for modeling plant development.  Precipitation (prcp) (mm), which affects soil moisture and crop water availability. Day length (dayl) (seconds) and shortwave radiation (srad) (W/m²), essential for estimating photosynthesis potential. Vapor pressure (vp) (Pa), which influences evapotranspiration and atmospheric demand.  Snow water equivalent (snow) (kg/m²), relevant for regions where snowmelt contributes to soil moisture. Potential evapotranspiration (pet) (mm), an estimate of atmospheric water demand based on climatic conditions.

\subsection{Soil Data}
Soil properties play a crucial role in determining crop productivity, water availability, and overall land suitability for agriculture. To incorporate these factors into our analysis, we utilize soil data from the Soil Survey Geographic Database (SSURGO), a high-resolution dataset developed by the Natural Resources Conservation Service (NRCS) of the U.S. Department of Agriculture (USDA) \citep{SSURGO}. SSURGO provides detailed soil property and classification data across the United States, enabling fine-scale agricultural and environmental assessments.

SSURGO provides detailed soil information at the field scale, with mapping resolutions ranging from 1:12,000 to 1:63,360. In this study, we extract soil attributes corresponding to cropland areas, ensuring alignment with CDL, OpenET, and DayMet datasets. To enhance spatial precision, we digitized soil information into a 30×30 m spatial resolution raster, enabling fine-scale analysis of soil properties relevant to agricultural productivity. The following five soil variables were selected based on their critical role in water retention, drainage, and overall land suitability:
    \begin{itemize}
        \item Available Water Storage (aws0100wta) (cm) – Represents the soil’s capacity to retain water for plant use within the top 100 cm.
        \item Slope Gradient (slopegraddcp) (\%) – Influences water runoff, erosion potential, and mechanized farming feasibility.
        \item Available Water Supply (awmmfpwwta) (cm) – Measures water availability within the most restrictive root zone, crucial under drought conditions.
        \item Drainage Class (drclassdcd) – A categorical variable indicating soil drainage potential, ranging from well-drained to poorly drained. We digitized this variable using a numeric scale, where "Excessively drained" is assigned 5.0, "Well drained" is 4.0, down to "Very poorly drained", which is 0.0.
        \item Hydrologic Soil Group (hydgrpdcd) – Classifies soils based on infiltration rates and runoff potential, essential for irrigation and flood risk management. This variable was digitized into four classes, with "A" mapped to 0.0, "B" to 1.0, "C" to 2.0, and "D" to 3.0.
    \end{itemize}
By rasterizing categorical soil properties into a numerical format, our approach allows for seamless integration with other spatial datasets, ensuring high-resolution, data-driven agricultural assessments (sample of rasterized soil properties has been visualized in Figure~\ref{fig:soil}).

\section{Multi-Model Vision Transformers}
For county-based crop yield forecasting, a multi-modal learning approach is essential due to the availability of multiple data sources contributing to crop productivity. To develop a robust forecasting model, it is necessary to integrate various modalities, enabling the model to learn from different factors influencing yield.

\begin{figure}[ht]
\centering
\captionsetup{font=footnotesize}%
\includegraphics[width=8cm, height=5cm]{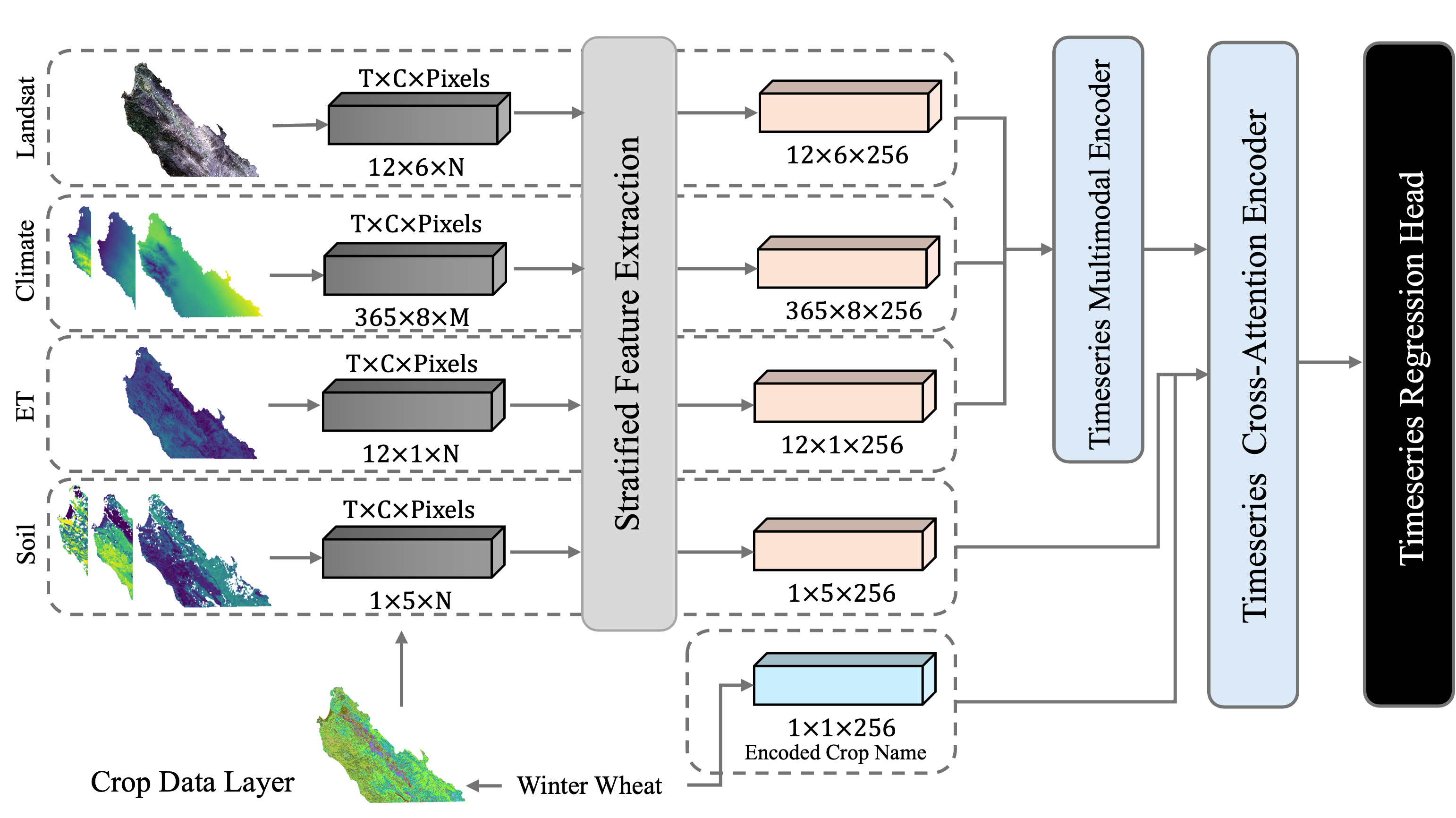} 
\caption{Multi-modal Vision Transformer model architecture. CDL is used to identify the crop type for pixel-level extraction across all variables and modalities. Due to differing spatial resolutions—particularly for climate data—the number of pixels (M vs. N) varies between modalities. Additionally, pixel counts differ by crop, so a stratified sampling strategy is applied to preserve the input distribution. To generalize the model across crops, the crop name is encoded as an additional input variable. A multi-modal Vision Transformer encoder is designed to learn feature representations at each time step for time-series yield prediction.
 }
\label{fig:model}
\end{figure}

In this study, the benchmark dataset includes several key inputs: satellite imagery from Landsat, which provides spatial and temporal insights into crop health and vegetation indices; climate data capturing temperature, precipitation, and other atmospheric conditions that influence crop growth; soil information representing soil type, moisture capacity, and fertility conditions across different regions; and evapotranspiration (ET), which is essential for understanding the water balance and irrigation demands. 

To accurately associate each crop with its respective location, the CDL is utilized for pixel-wise crop identification, ensuring that model predictions align with actual crop distribution at the county level. By leveraging CDL as a baseline, we can systematically extract and integrate multi-source data, creating a comprehensive dataset for crop-specific analysis.

The proposed architecture is designed to predict county-level crop yields by integrating heterogeneous spatiotemporal data through a stratified feature extraction pipeline and a multi-modal encoder. As illustrated in Figure~\ref{fig:model}, the model processes five primary data sources:
\begin{enumerate}
    \item Satellite Imagery (Landsat): Monthly Landsat data is structured as a sequence with a shape of $12 \times 6 \times N$, representing 12 time steps (months), 6 channels (e.g., reflectance bands or vegetation indices), and N pixels. These are passed through a feature extractor to produce embeddings of size $12 \times 6 \times 256$.
    \item Climate Data: Daily climate variables (e.g., temperature, precipitation, humidity) are captured across 365 time steps, 8 channels, and M pixels, resulting in an input of shape $365 \times 8 \times M$. After feature extraction, this becomes a $365 \times 8 \times 256$ representation.
    \item Evapotranspiration (ET): Monthly ET data is structured as $12 \times 1 \times N$, with one channel representing the ET value per pixel per month. It is transformed to a $12 \times 1 \times 256$ embedding.
    \item Soil Properties: Since soil characteristics (e.g., texture, pH, organic matter) are static and do not change over time, they are input as a single time step with five channels across N pixels, shaped as $1 \times 5 \times N$, and converted into a $1 \times 5 \times 256$ embedding.
    \item Crop Type: A categorical identifier for the crop is encoded and embedded as a constant input with shape $1 \times 1 \times 256$. This enables the model to generalize across multiple crops by explicitly incorporating crop identity into the learning process.
\end{enumerate}

Each of these modalities is independently passed through a Stratified Feature Extraction module, which standardizes the feature representations to a uniform embedding size of 256. These extracted features are then fused in a Multi-Modal Encoder, which jointly learns temporal and cross-modal interactions. Finally, a Time-Series Regression Head generates the predicted yield output for the specified crop and region.

This architecture allows the model to dynamically track crop development throughout the growing season, while also accounting for spatially fixed inputs like soil and crop type. By unifying satellite, climate, ET, soil, and categorical crop information within a consistent spatiotemporal framework, the model improves both the accuracy and generalizability of county-level yield forecasting. 

\subsection{Experimental Setting}
The model consists of 8 transformer layers, each with 6 attention heads and an embedding dimension of 256. The dataset is split temporally, reserving the years 2021 and 2022 for testing, while the remaining years are used for training and validation through 5-fold cross-validation. The training process uses a fixed learning rate of 0.0001 and a weight decay of 0.01, optimized using the AdamW optimizer with a mean squared error (MSE) loss function.

\begin{figure*}[ht]
\centering
\captionsetup{font=footnotesize}%
\includegraphics[width=15cm, height=4.5cm]{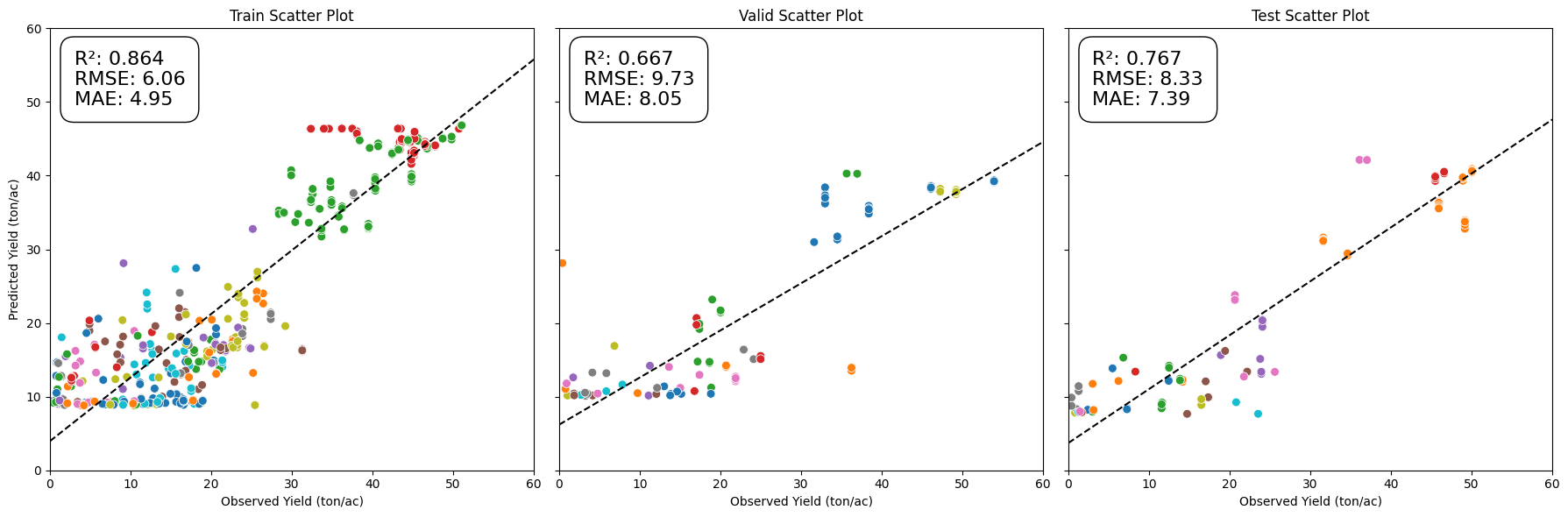} 
\caption{Model Performance on Multi-Crop Yield Prediction (2008–2022)
Scatter plots illustrate the performance of a multimodal model in predicting crop yields across 70 crop types over 14 years. The model was trained on data from 2008 to 2018, validated on 2019–2020, and tested on 2021–2022. Each point represents a field observation, with colors denoting different crop types. The dashed line represents the ideal 1:1 relationship between observed and predicted yields. 
 }
\label{fig:results}
\end{figure*}

\begin{table}[ht]
\centering
\footnotesize
\caption{Model performance metrics (RMSE, and MAE (ton/ac)) for some selected crops with more availability across training, validation, and test datasets}
\begin{tabular}{p{1.5cm}|cc|cc|cc}
\toprule
\multirow{2}{*}{Crop} & \multicolumn{2}{c|}{Train} & \multicolumn{2}{c|}{Validation} & \multicolumn{2}{c}{Test} \\
                     & RMSE & MAE & RMSE & MAE &  RMSE & MAE \\
\midrule
Corn & 6.17 & 5.40  & 8.13 & 8.10 & 10.10 & 9.98 \\
Tomatoes & 4.17 & 3.54 & 7.69 & 6.38 & 6.13 & 5.75 \\
Almond & 7.85 & 6.91 & 6.5 & 5.98 & 8.41 & 8.36 \\
Alfalfa & 7.36 & 6.66 & 7.55 & 7.50 & 6.74 & 6.56 \\
Grapes & 3.13 & 2.61 & 2.69 & 2.36 & 3.24 &  3.10\\
Strawberries & 6.7 & 5.94 & 4.18 & 4.14 & 5.47 & 5.45 \\
Walnuts & 5.21 & 4.96 & 7.44 & 7.23 & 8.36 & 7.21 \\
Greens & 3.76 & 3.12 & 4.18 & 4.10 & 3.47 & 2.66 \\
\bottomrule
\end{tabular}
\label{tab:crops}
\end{table}

\section{Results and Discussion}
To evaluate the performance of the proposed model, we employed three widely used regression metrics: the coefficient of determination (R²), root mean squared error (RMSE), and mean absolute error (MAE). These metrics were computed across all crops to assess the model’s accuracy and generalizability. Since the model performs time-series predictions throughout the crop growing season, the results reflect yield estimation dynamics tailored to the specific phenological timelines of each crop. This section presents a comprehensive analysis of the model’s performance, highlighting its strengths across diverse crop types, seasonal patterns, and varying environmental conditions in California.

Shown in Figure~\ref{fig:results}, with a strong R² of 0.864 on the training set (2008–2018), the model effectively captures complex patterns in the data. While performance naturally decreases on unseen data, the validation set (2019–2020) still shows a reasonable R² of 0.667, and the test set (2021–2022) achieves an improved R² of 0.767, indicating good generalization capability. The relatively low RMSE and MAE across all splits suggest the model maintains accurate predictions even as crop types and environmental conditions vary across years. The consistent spread of crop-specific predictions, visualized by color-coded dots, further highlights the model’s robustness across different crop categories. 

While the model demonstrates strong overall performance, it exhibits variability across different crop types, highlighting both its strengths and limitations shown in Table~\ref{tab:crops}. For instance, the model achieves relatively low RMSE and MAE values for crops such as Grapes, Greens, and Tomatoes, indicating high predictive accuracy and consistency. These crops benefit from more stable patterns or possibly richer data availability. However, the model struggles with others like Corn, Almond, and Walnuts, particularly in the test set where RMSE values exceed 8–10 ton/ac, suggesting limited generalization for crops with higher variability or fewer samples. Additionally, while Strawberries show improved performance in the validation set, they experience a slight decline in test accuracy, pointing to inconsistency in certain cases. These results indicate the need for longer and more comprehensive time series data to better capture seasonal trends, management practices, and climate variability. Future work should focus on expanding the temporal depth and spatial coverage of the dataset, incorporating crop-specific features, and exploring ensemble or crop-specific modeling approaches to improve performance across all crop types.

\section{Conclusion}
In this study, we introduced a comprehensive benchmark dataset and a multi-modal deep learning framework for county-level crop yield forecasting across California. By integrating diverse data modalities—including Landsat imagery, climate variables, evapotranspiration, and soil properties—from 2008 to 2022, we developed a robust model capable of capturing complex spatial and temporal dynamics in crop productivity. The proposed model achieved an overall R² of 0.76 across more than 70 crop types, demonstrating its effectiveness in generalizing across different crop groups and environmental conditions.

While the focus of this work is on county-level yield prediction, a key limitation is the absence of field-level ground-truth yield data, which would allow for fine-grained, precision agriculture applications. However, the model and benchmark dataset are designed to be generalizable and trainable at finer spatial resolutions. If future researchers or practitioners have access to field-level yield measurements, this framework can be readily adapted for field-based crop yield forecasting, providing a scalable solution for high-resolution agricultural monitoring.

Overall, this work provides a valuable foundation for the remote sensing, agricultural, and machine learning communities, enabling further research in yield prediction, climate resilience, and sustainable farming practices. The dataset and model code are made publicly available to promote transparency, reproducibility, and collaborative advancement in the field.
{
    \small
    \bibliographystyle{ieeenat_fullname}
    \bibliography{main}
}

\end{document}